\documentclass[runningheads]{llncs}

\usepackage{eccv}

\usepackage{eccvabbrv}

\usepackage{graphicx}
\usepackage{booktabs}

\usepackage[accsupp]{axessibility}  

\usepackage{hyperref}

\definecolor{my_red}{rgb}{0.835, 0.309, 0.243}
\definecolor{my_green}{rgb}{0.333, 0.635, 0.345}
\definecolor{my_blue}{rgb}{0.298, 0.505, 0.913}

\newlength\savewidth
  
\usepackage{orcidlink}
\usepackage{multirow}
\usepackage{caption}
\usepackage{bm}

\begin{document}

\title{Approaching Outside: Scaling Unsupervised 3D Object Detection from 2D Scene} 

\titlerunning{Scaling Unsupervised 3D Object Detection from 2D Scene}

\author{
Ruiyang Zhang\inst{1}\orcidlink{0009-0009-8397-5447}
\and
Hu Zhang\inst{2}\orcidlink{0009-0009-9892-9515}
\and
Hang Yu\inst{3}\orcidlink{0000-0003-3444-9992}
\and
Zhedong Zheng\inst{1}\thanks{Corresponding author.}\orcidlink{0000-0002-2434-9050} 
}

\authorrunning{R.~Zhang \etal}

\institute{
FST and ICI, University of Macau, China \\
\and
CSIRO Data61, Australia \\
\and
Shanghai University, China \\
\email{ruiyang.061x@gmail.com, Hu1.Zhang@csiro.au, yuhang@shu.edu.cn, zhedongzheng@um.edu.mo} \\
\url{https://github.com/Ruiyang-061X/LiSe} 
}

\maketitle

\begin{abstract}

The unsupervised 3D object detection is to accurately detect objects in unstructured environments with no explicit supervisory signals.
This task, given sparse LiDAR point clouds, often results in compromised performance for detecting distant or small objects due to the inherent sparsity and limited spatial resolution. In this paper, we are among the early attempts to integrate LiDAR data with 2D images for unsupervised 3D detection and introduce a new method, dubbed LiDAR-2D Self-paced Learning (LiSe). We argue that RGB images serve as a valuable complement to LiDAR data, offering precise 2D localization cues, particularly when scarce LiDAR points are available for certain objects. Considering the unique characteristics of both modalities, our framework devises a self-paced learning pipeline that incorporates adaptive sampling and weak model aggregation strategies. The adaptive sampling strategy dynamically tunes the distribution of pseudo labels during training, countering the tendency of models to overfit easily detected samples, such as nearby and large-sized objects. By doing so, it ensures a balanced learning trajectory across varying object scales and distances. The weak model aggregation component consolidates the strengths of models trained under different pseudo label distributions, culminating in a robust and powerful final model. Experimental evaluations validate the efficacy of our proposed LiSe method, manifesting significant improvements of $+7.1$\% AP$_{BEV}$ and $+3.4$\% AP$_{3D}$ on nuScenes, and $+8.3$\% AP$_{BEV}$ and $+7.4$\% AP$_{3D}$ on Lyft compared to existing techniques.


\end{abstract}

\section{Introduction}

\begin{figure*}[t]
    \centering
    \includegraphics[width=\linewidth]{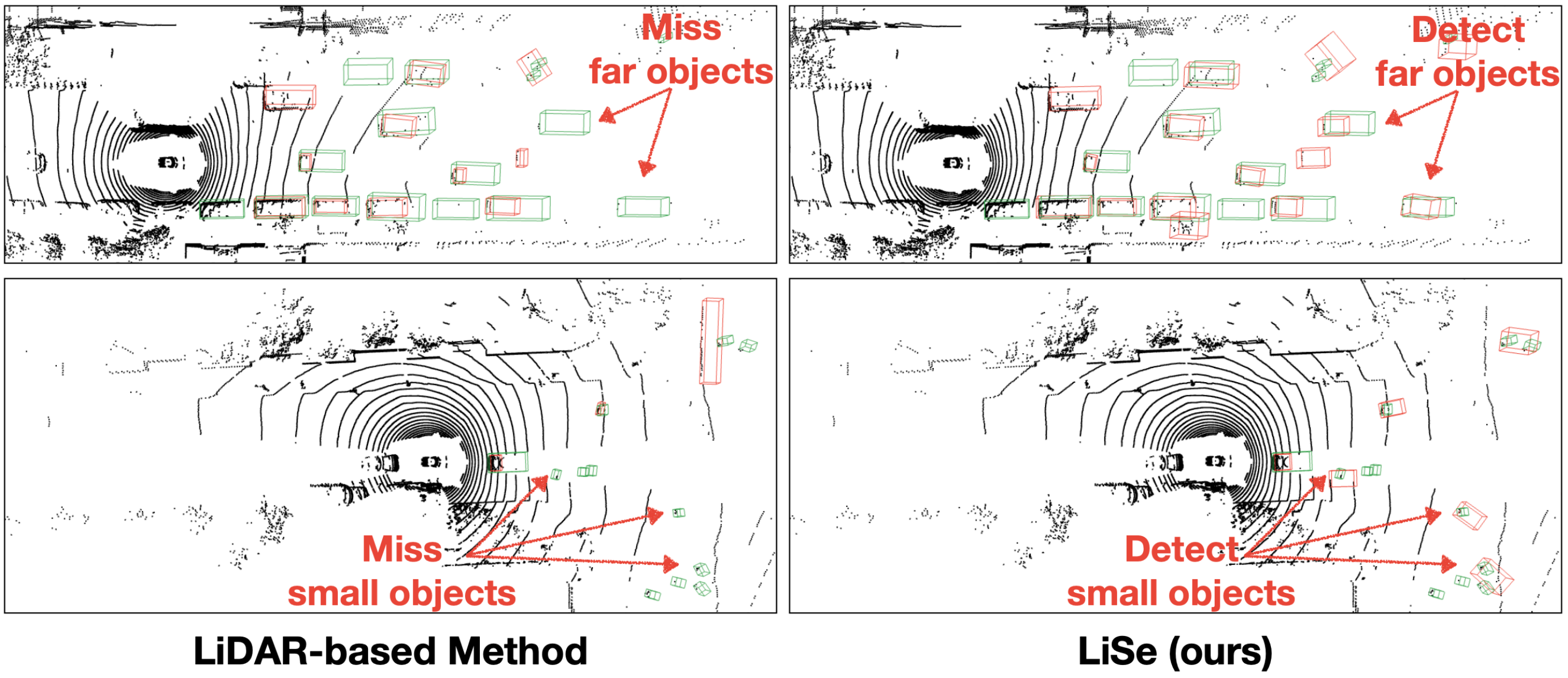}
    \caption{
    We show typical limitations of LiDAR-based methods for unsupervised 3D object detection. Compared with the prevailing LiDAR-based method, \ie, MODEST~\cite{you2022learning} generally misses objects in the distance and small objects (left), our proposed method LiSe successfully recalls such objects (right). Best viewed in color: the \textcolor{my_green}{green} boxes are ground truth labels and the \textcolor{my_red}{red} boxes are predictions.
    }
    \label{fig:drawback}
\end{figure*}

Unsupervised 3D object detection in the context of autonomous driving aims to discover potential 3D objects in an unsupervised manner~\cite{you2022learning, zhang2023towards, najibi2022motion}. 
The key underpinning unsupervised 3D detection is to develop intelligent algorithms that can effectively reason about and adapt to the vast array of potential object classes and their various manifestations in real-world scenarios without explicit prior knowledge or labeled data. The process involves not only accurately estimating the 3D position of objects but also learning to identify previously unseen object types and handle unpredictable environmental conditions, which remains challenging to 2D-based methods~\cite{guo2024dsca,chen2023pipa}. This technology is crucial for ensuring the safety and efficiency of autonomous vehicles as they navigate through complex and unpredictable road environments. The ability to accurately detect 3D objects allows these systems to anticipate potential hazards, make informed decisions, and react accordingly. It can be widely applied to real-world applications, including pedestrian protection~\cite{choi2016safety, gandhi2007pedestrian}, auto-driving assistance systems~\cite{hu2023planning, jia2023driveadapter, huang2023gameformer, horgan2015vision} and traffic management~\cite{shi2023open, liu2021smart}. The inherent challenge lies in designing models capable of extracting discriminative features from sparse and noisy sensor data, such as point clouds and images, while simultaneously overcoming issues related to class imbalance, scale variation, and partial occlusions.

Most existing works usually focus on mining the LiDAR data to discover unlabeled 3D objects~\cite{you2022learning, zhang2023towards, najibi2022motion}. For instance, some works~\cite{you2022learning, zhang2023towards} utilize the rule-based generation of pseudo boxes followed by a self-training process, and another line of works~\cite{najibi2022motion} harness the motion cues provided by LiDAR scene flow to identify potential dynamic objects. While LiDAR data provides accurate depth information and a comprehensive perception of the surrounding environment, it is limited by the inherent sparsity and spatial resolution. This sparsity challenge becomes particularly pronounced in scenarios involving long-range or small-scale objects (see Figure~\ref{fig:drawback}), where the dearth of LiDAR points returned significantly compromises the discriminative power required to accurately segregate foreground entities from their background context. Current methodologies often adopt an iterative training pipeline that commences with initial object proposals followed by pseudo-label refinement. However, overdependence on LiDAR alone can lead to a blind spot for detecting diminutive or distant objects, thus culminating in suboptimal overall detection capabilities.

In this paper, we thus propose a novel LiDAR-2D Self-paced Learning ~(\textbf{LiSe}) for unsupervised 3D detection, which integrates LiDAR data with 2D images. It aims to leverage the rich textual and RGB color information in 2D scenes to overcome the limitations of LiDAR in detecting distant and small objects. We adopt the off-the-self multi-traversal method for LiDAR-based 3D detection~\cite{you2022learning}, while applying the 2D detection~\cite{liu2023grounding} and segmentation~\cite{kirillov2023segment} with 3D lifting for image-based 3D detection. We observe that two modalities are complementary on the objects with different distances and resolution, and serve as good initialization seed. Then we apply the self-paced training strategy to propagate the object label and refine the box prediction. During training, we observe that the models tend to overfit to the common category, \eg, cars, and gradually lose the ability to detect relatively rare objects, \eg, bicycles. To alleviate the diminishing detection ability of such long-tail samples, we introduce an adaptive sampling strategy that dynamically adjusts the distribution of training data based on the feedback of the model. Therefore, we could obtain the snapshots trained with different data distributions during the training process, and thus the learned snapshots are inherently with complementary focuses. We further propose the weak model aggregation strategy to merge all snapshot weights along the self-paced learning process as the final model. We conduct extensive quantitative experiments and qualitative analyses to validate the effectiveness of our method. In conclusion, our contributions are summarized as follows:
\begin{itemize}
    \item Considering the inherent sparsity of LiDAR data, we propose a new approach, called \textbf{LiSe}, jointly leveraging 2D images and 3D LiDAR to improve the pseudo label quality in all ranges. The rich texture in 2D images provides a straightforward discovery of small and distant objects. 
    \item Considering the imbalanced object distribution in self-training, we propose an adaptive sampling strategy to explicitly emphasize the long-tailed objects, followed by the weak model aggregation, which iteratively fuses the strengths of different snapshots into a final stable model.
    \item Extensive experiments on nuScenes and Lyft verify the effectiveness of the proposed method, surpassing state-of-the-art by a clear margin on both AP$_{3D}$ and AP$_{BEV}$. Especially, for long-range detection (50-80m), the AP$_{BEV}$ metric even exceeds that of fully supervised model.
\end{itemize}

\section{Related Work}

\noindent \textbf{Unsupervised 3D Object Detection.} Unsupervised 3D object detection is attracting increasing interest within the research community~\cite{tian2021unsupervised, wong2020identifying, dewan2016motion, zhang2023opensight}. Cen~\etal~\cite{cen2021open} utilize a fully-supervised detector to generate proposals for unknown classes. However, this approach struggles with overconfidence issues and lacks the ability to generate proposals for semantically distinct classes. Studies like MODEST~\cite{you2022learning} and OYSTER~\cite{zhang2023towards} have explored a self-training pipeline to incrementally discover more objects from LiDAR data. Nevertheless, the inherent sparsity of LiDAR data hampers their ability to detect small objects. Najibi~\etal~\cite{najibi2022motion} employ scene flow to identify objects in motion, training detection model with generated pseudo boxes. Najibi~\etal~\cite{najibi2023unsupervised} propose distilling VLM knowledge into 3D detection model, addressing detection task by initially segmenting based on text references, followed by 3D box fitting. Different from existing works, our work aims to leverage the information density in 2D scenes to enhance recognition of distant and small objects. We also develop adaptive sampling strategy to address data distribution imbalance problem during self-training process.

\noindent \textbf{Open-vocabulary 2D Detection.} Open-vocabulary 2D detection methods can be categorized into two groups: (1) Knowledge distillation methods~\cite{bangalath2022bridging, wang2023object, cho2023open} focus on transferring the extensive open-vocabulary knowledge from VLMs into closed-set 2D detectors. Therefore, the detection capability is limited by the scope of the teacher VLM. 
(2) Region-text pretraining methods~\cite{buettner2023enhancing, li2022grounded, zhang2022glipv2} emphasize learning from region captions through pretraining at the region level, albeit with high computational costs due to the large scale of pretraining datasets. Similarly, our work leverages pretrained open-vocabulary detection models to recognize objects in 2D images. It helps us to instill 2D prior to the 3D detection. We also consider to eliminate the negative impact of the class imbalance and overfitting during the self-training process.  

\noindent \textbf{Image and LiDAR Fusion.} In the realm of closed-set object detection, recent works have begun to study image and LiDAR data fusion. These works can be divided into two categories, \ie, sequential fusion~\cite{qi2018frustum, wang2019frustum} and parallel fusion~\cite{chen2017multi, pang2020clocs}. (1) Sequential based approaches~\cite{qi2018frustum, wang2019frustum} use 2D base models to initiate the 3D detection pipeline. The drawback here is that failures in 2D models accumulate in the 3D detection model, which inevitably degrades the overall performance. (2) Parallel fusion approaches integrate two modalities at different stages of the pipeline, including early fusion at the input stage~\cite{wei2021fgr}, deep fusion at the feature stage~\cite{chen2017multi, ku2018joint, liang2018deep, guo2024cdk}, and late fusion at the output stage~\cite{pang2020clocs,zheng2018pedestrian}. A primary challenge in parallel fusion is to ensure semantic alignment, due to substantial gap between characteristics of images and point clouds. Different from existing works, our method fuses two modalities by combining box seeds from both. We introduce self-paced learning to progressively update the box labels, offering robustness where the failure of 2D models does not halt the pipeline.

\noindent \textbf{Self-paced Learning.} Self-paced learning~\cite{kumar2010self} describes a self-directed learning process where the distribution of training data is dynamically adjusted based on the model performance. This concept has seen broad application in computer vision field, including image classification~\cite{tang2012self, zhang2021flexmatch, yang2020curriculum}, object detection and localization~\cite{soviany2021curriculum, shrivastava2016training, zhang2019leveraging}, scene segmentation~\cite{li2017multiple, sakaridis2019guided}, and video processing~\cite{zhang2017spftn, liang2016learning}. However, these strategies are predominantly focused on the image and video domains. Diverging from existing works, we have tailored a self-paced learning strategy specifically for the 3D unsupervised detection task, incorporating unique 3D attributes such as object distance and volume information.

\section{Method}
\begin{figure*}[t]
    \centering
    \includegraphics[width=\linewidth]{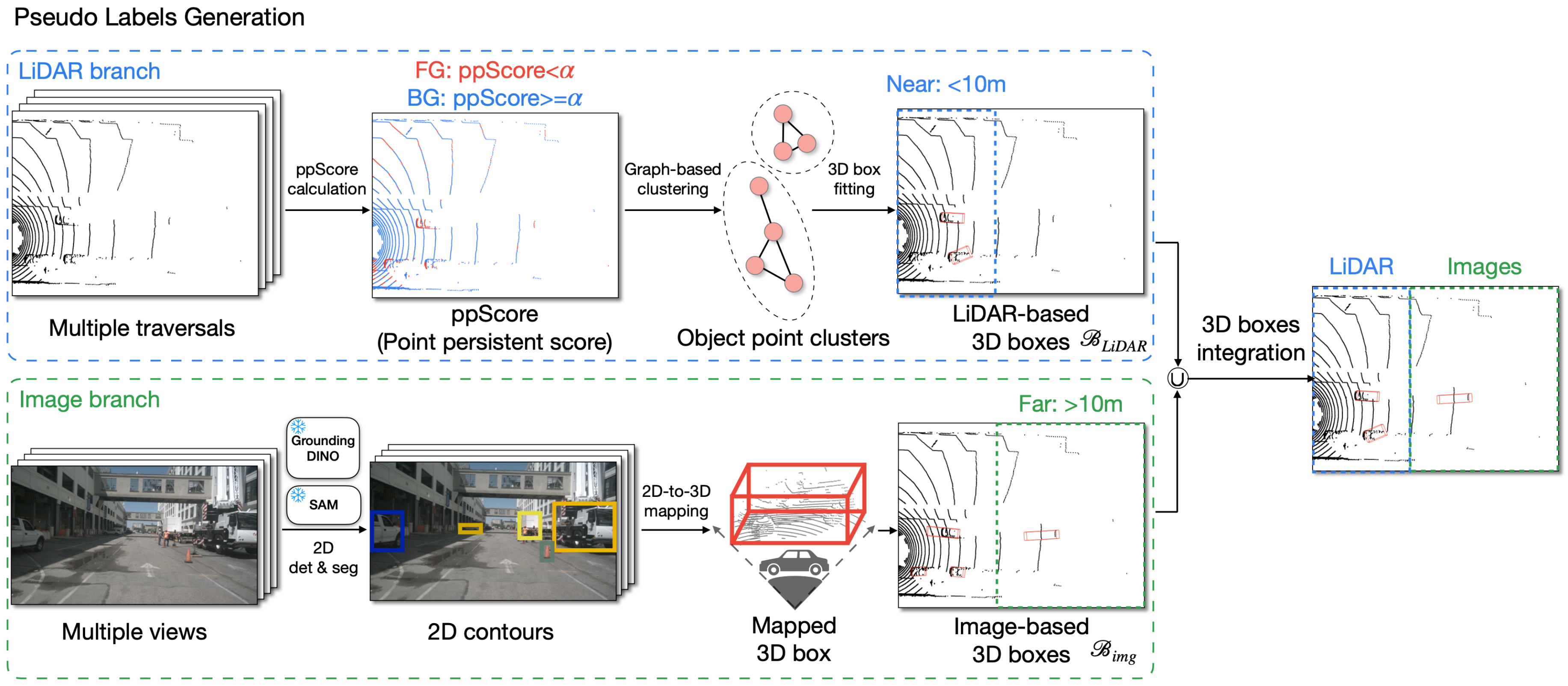}
    \caption{
    Illustration of the pseudo label generation process in LiSe, which distinctively harnesses information density from 2D scenes to complement LiDAR data. Our approach involves a generation method tailored for each modality to obtain LiDAR-based and image-based 3D boxes. In the LiDAR branch, an off-the-shelf multi-traversal based method generates pseudo labels, primarily covering near-range objects. Concurrently, the image branch uses a pretrained open-vocabulary 2D detector and Segment-Anything-Model to generate 2D contours from images, which are then mapped into 3D space. Following this, a distance-aware 3D boxes integration process fuses boxes from both LiDAR and image modalities. Notably, image-based boxes at longer ranges (\eg, > 10m) are merged with LiDAR-based 3D boxes. This integration addresses the limitations of LiDAR-based method in detecting long-range and small objects. The resulting pseudo labels are proficient in originally challenging samples (\eg, distant and small objects) for LiDAR-based methods, laying a solid foundation for enhancing detection model performance on these challenging cases.
    }
    \label{fig:structure_part_1}
\end{figure*}

We present a detailed description of our method in this section and structure it into three parts: (1) integration of LiDAR data with 2D scene (see Figure~\ref{fig:structure_part_1}), (2) adaptive sampling strategy (see Figure~\ref{fig:structure_part_2}), and (3) weak model aggregation (see Figure~\ref{fig:structure_part_2}).

\subsection{Integration of LiDAR Data with 2D Scene}

\noindent \textbf{Pseudo-boxes from LiDAR.} In our work, we apply a multi-traversal approach to extract significant objects from LiDAR data. Multi-traversal approach is based on the idea that when a vehicle traverses the same location multiple times, entities that remain unchanged in both position and state are likely static background elements, \eg buildings. Conversely, items that shift in location are probable foreground objects, \eg moving cars. Specifically, we first conduct the data processing for given LiDAR data. For locations visited more than once, the LiDAR scans collected at these locations are combined. We then use the GPS/INS data which provides accurate information on vehicle location and the rotation matrix to calibrate the data so that different LiDAR scans are aligned into same coordinate system. After the alignment, we calculate the point persistency score (ppScore)~\cite{you2022learning} of each point $\tau(u)$ to quantify whether it belongs to unchanging or changing objects. The higher ppScore indicates a more static point and lower ppScore indicates a more dynamic one. 

With the calculated ppScore, a clustering process that considers both the similarity of ppScore and the actual geometric distance between points is utilized to segment the entire point cloud into distinct clusters. A graph is constructed where points within radius threshold $r_t$ of each other are connected by an edge, and the weight of this edge is calculated as the absolute difference of their ppScores $\lvert\tau(u) - \tau(v)\rvert$. Following graph construction, a variant of the DBSCAN~\cite{ester1996density} algorithm which is adapted for application on the graph is used on the constructed graph, resulting in numerous clusters of points with similar ppScores and proximate geometric distances. A filtering process that excludes clusters where the top $K$ percent of points with ppScore above threshold $\alpha$ is applied, designating these as static clusters (\eg, large building walls). The remaining clusters are treated as foreground objects. Finally, an off-the-shelf bounding box fitting algorithm~\cite{zhang2017efficient} is applied to each cluster to create a 3D box. 

\noindent \textbf{Pseudo-boxes from Images.} For 3D pseudo-boxes generation from images, we employ an off-the-shelf open-vocabulary 2D detector, \eg, GroundingDINO~\cite{liu2023grounding} to first identify discriminative objects within the images. To realize this purpose, we construct the detection prompt by concatenating discriminative class names together, feeding it into GroundingDINO, and obtaining a collection of 2D boxes. Usually, the detected 2D boxes often contain substantial background areas, which do not accurately reflect the shapes of real-world objects. Direct employment of these 2D boxes for subsequent processing can result in imprecise 3D box estimations. Fortunately, the 2D boxes naturally serve as initial prompts for the Segment-Anything-Model (SAM)~\cite{kirillov2023segment}. By inputting these 2D boxes into SAM as prompts, we instead obtain refined 2D masks. These masks reflect the actual contours of the targeted objects, significantly mitigating the drawbacks inherent in utilizing 2D boxes directly. To estimate 3D boxes from images, a projection process from 3D to 2D is then applied and can be formulated as:
\begin{equation}
    \hat{u}_{i} = \textbf{K} \cdot \textbf{E} \cdot u_{i}, \ (i = 1,...,m),
\end{equation}
where $u_i$ is one LiDAR point in the 3D space, $\textbf{K}$ is the intrinsic matrix, $\textbf{E}$ is the extrinsic matrix, and $m$ is number of points in point cloud. For those projected 2D points lying in the masks, we reserve their corresponding 3D points. This process is equal to building a frustum extending from the ego center of the vehicle to 2D mask. This frustum is considered to contain the point cluster corresponding to the detected object. We then apply the region growth algorithm~\cite{adams1994seeded} to get the cluster with the most points. Subsequently, tight external 3D bounding box is estimated based on the cluster. All the generated 3D boxes are then consolidated to form the final pseudo labels for a single LiDAR point cloud. Ultimately, we achieve a comprehensive set of pseudo labels for the training dataset, entirely independent of any ground-truth 3D annotations.

It is worth noting that due to the rich texture information in 2D images and the strong detection ability of the employed open-vocabulary 2D detector, many distant and small objects, which are usually challenging to identify in LiDAR data, can be recognized. The pseudo labels derived from images can thus serve as a robust complement to those obtained from LiDAR, potentially enhancing the overall quality and coverage of the training data.

\noindent \textbf{Integration between LiDAR and 2D scene.} To enhance the integration of pseudo boxes from both LiDAR and images for training models, we employ a distance-aware strategy. This approach optimally leverages the complementary characteristics of two data sources. We begin by establishing a predefined range, and then we selectively include image-derived boxes that lie within this range, alongside LiDAR-generated boxes. Final bounding boxes can be derived from:
\begin{equation}
    \mathcal{B}_{final} = \mathcal{B}_{LiDAR} \cup \{b_i \mid d(b_i) \geq d_{min}, b_i \in \mathcal{B}_{img} \}, 
\end{equation}
where $\mathcal{B}_{img}$ denotes all image-derived boxes and $\mathcal{B}_{LiDAR}$ denotes LiDAR-generated boxes. $b_i$ is one image-derived 3D box, $d(b_i)$ is the distance between box $b_i$ and the ego car. $d_{min}$ is the determined range value. Considering that objects in close proximity typically exhibit a high density of LiDAR points, LiDAR data alone is often sufficient for precise estimations. Our distance-aware strategy allows for flexible exclusion of image-derived boxes in these near-range areas by adjusting range values. It helps to prevent possible conflicts with LiDAR-generated boxes.

\subsection{Adaptive Sampling Strategy}

\begin{figure*}[t]
    \centering
    \includegraphics[width=\linewidth]{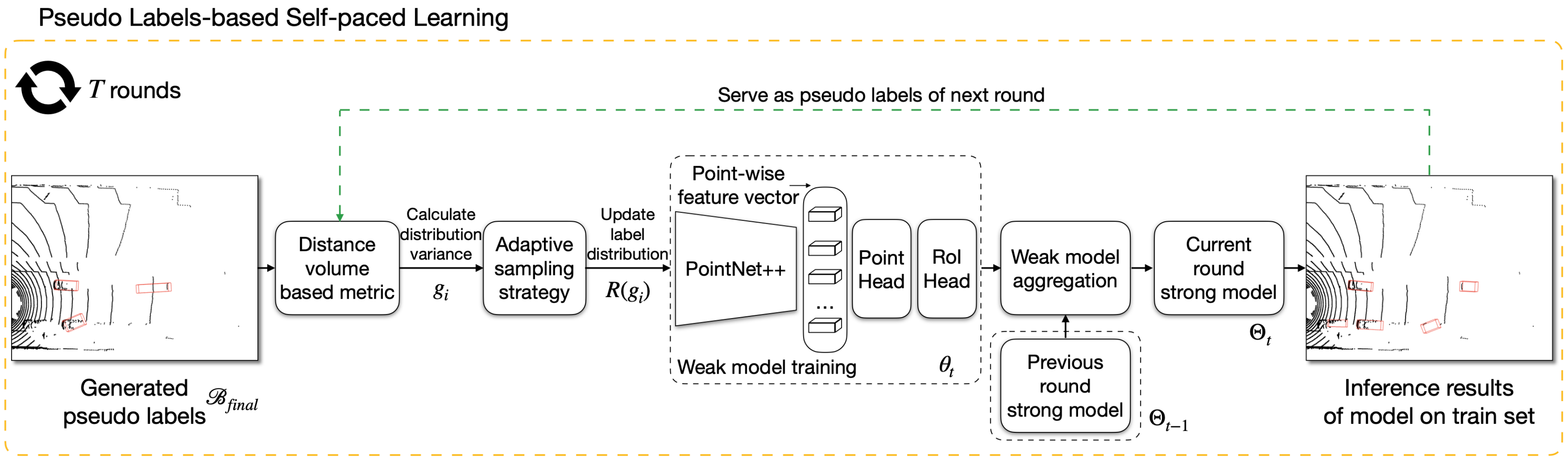}
    \caption{
    Illustration of the self-paced learning process in LiSe. Initial distribution of objects and inference distribution after training are first calculated with the distance volume-based metric. Adaptive sampling strategy thus updates sampling rates for different objects based on changes in two distributions. We further consider weak model aggregation to combine newly trained model with previously aggregated model to obtain a stronger, more robust model for current round. Finally, we iteratively update distribution of pseudo labels and model weight for $T$ rounds to obtain the final model.
    }
    \label{fig:structure_part_2}
\end{figure*}

Despite integrating 2D scenes into 3D pseudo-boxes is able to recall the missed distant and small objects, the model tends to be biased toward easier samples, \eg, closer or larger objects in training. Such bias persists throughout all training rounds and the reason behind is attributed to the imbalanced data distribution. We thus propose an adaptive sampling strategy, dynamically balancing different objects throughout the training phases (see Figure~\ref{fig:structure_part_2}).

We first propose distance volume-based metric, which leverages general properties in 3D world, \ie, distance and volume to categorize objects. For distance-based categorization, we adopt the criteria in MODEST~\cite{you2022learning}, dividing objects into near- and far-range ones: objects within 0-30m are considered as near objects and those beyond 30m are categorized as far objects. For volume-based categorization, we consult GPT-4 for general information on the volume and size of common categories. We then classify objects with a volume smaller than \(5m^3\) as small objects, and those larger than \(5m^3\) as large objects. For example,  common categories such as pedestrians, cyclists will be categorized as small objects, and typical cars or other vehicles will be attributed to large ones.

Based on the distance volume-based metric, we calculate the initial object distribution before training and inference distribution after training. We analyze the differences between two distributions: For object groups whose probability in inference distribution is significantly increased, we adaptively downsample these objects in the next round. Conversely, for object groups whose probabilities are decreased in the inference, we adaptively upsample these groups accordingly. Therefore, we introduce a sampling score, which can be formulated as:
\begin{equation}
R(g_i) =
\begin{cases}
1 - (Q(g_i) - Q_{init}(g_i)) & \text{if } Q(g_i) > Q_{init}(g_i) \\
1 + (Q_{init}(g_i) - Q(g_i)) & \text{if } Q(g_i) \le Q_{init}(g_i)
\end{cases},
\end{equation}
where $g_i$ is one type of objects grouped by distance volume-based metric. $Q(g_i)$ is sampling probability in inference distribution and $Q_{init}(g_i)$ is sampling probability in initial distribution. $R(g_i)$ is new sampling score for objects in group $g_i$ in next round. Adaptive resampling can calibrate model towards harder samples and away from easier ones, thus resulting in self-paced learning process.

\subsection{Weak Model Aggregation}

The models obtained in different rounds tend to be proficient in different object groups, with the adaptive sampling strategy assigning varied sampling ratios. For example, while a model trained in the $t$-th round excels at identifying large objects, the model in the $(t + 1)$-th round takes more attention to detecting small objects with the increased sampling rate for small objects. The models obtained in different rounds have their unique bias and lack a comprehensive detection ability. Therefore, we refer to these models as ``weak models'', and introduce weak model aggregation, which combines these weak models to create a robust, stronger model (see Figure~\ref{fig:structure_part_2}). We select a model as initialization starting from round $T_{s}$. Similar to weight-average approaches~\cite{zheng2022adaptive,tarvainen2017mean}, we average each weak model in subsequent rounds with the previous aggregated strong model, and the obtained model serves as the strong one for the current round. An aggregation coefficient $\lambda$ is utilized to balance the influence of the previous strong model and the current weak model. The calculation process can be formulated as:
\begin{equation}
\Theta_{t} =
\begin{cases}
\theta_{t} & \text{if } 1 \le t < T_{s} \\
\lambda \cdot \Theta_{t-1} + (1 - \lambda) \cdot \theta_{t} & \text{if } 
T_{s} \le t \le T \\
\end{cases},
\end{equation}
where $t$ is round number, $T_{s}$ is start round to perform weak model aggregation, and $T$ is total number of self-paced learning rounds. $\lambda$ is aggregation coefficient. $\theta_{t}$ is current weak model in $t$-th round, $\Theta_{t-1}$ is aggregated strong model obtained in $(t-1)$-th round, and $\Theta_{t}$ is the aggregated strong model in $t$-th round.

\subsection{Pseudo Labels-based Self-paced Learning}

We unify integrated pseudo labels, adaptive sampling strategy, and weak model aggregation into a self-paced learning process (see Fig.~\ref{fig:structure_part_2}). Specifically, it consists of two stages: seed training and self-training. In seed training, integrated pseudo labels $\mathcal{B}_{final}$ are used to train an initial detector $\Theta_{0}$. Self-training is an iterative process repeated for $T$ rounds. In $t$-th round, detector trained from previous round $\Theta_{t-1}$ first conducts inference on the training set to obtain pseudo training labels for the current round. The pseudo training labels are then redistributed with our proposed adaptive sampling strategy to counter the bias towards easier object groups, \eg, near-range and large objects. Then the updated pseudo labels are harnessed to train a new detector $\theta_{t}$. Weak model aggregation aggregates the weak model in current round $\theta_{t}$ and strong model from previous round $\Theta_{t-1}$ into a strong model for current round $\Theta_{t}$. Different from vanilla self-training, in our process, distribution of pseudo training labels is adjusted based on model feedback, which results in a self-paced learning process.

\section{Experiment}

\noindent \textbf{Dataset.} We conduct unsupervised 3D detection experiments on nuScenes~\cite{caesar2020nuscenes} and Lyft~\cite{houston2021one}, two widely recognized benchmarks in autonomous driving. In our experiment, we follow the basic dataset configuration in MODEST~\cite{you2022learning}. Specifically, we only utilize LiDAR point clouds which are collected from locations with more than one sample in order to satisfy the multi-traversal requirement. For nuScenes, the final used data consists of 3,985 training keyframes and 2,412 testing ones. For Lyft, we use 11,873 training samples and 4,901 testing samples. We emphasize that the ground-truth 3D annotations are not used in our training and they are just involved in the testing to evaluate model performance.

\noindent \textbf{Metric.} Two different metrics AP$_{BEV}$ and AP$_{3D}$ are considered. AP$_{BEV}$ focuses on accuracy from the Bird's Eye View (BEV), while AP$_{3D}$ considers additional height information and evaluates detection results in 3D space, thus offering more comprehensive assessment. Furthermore, we consider objects \emph{w.r.t.} the distance and report evaluation results for objects in the near range (0-30m), middle range (30-50m), far range (50-80m), and the full range (0-80m).

\noindent \textbf{Implementation Details.} In the training, we adopt PointRCNN~\cite{shi2019pointrcnn} as the backbone. In each self-paced training round, we train the model for 80 epochs on nuScenes and 60 epochs on Lyft. We adopt AdamOneCycle~\cite{smith2017cyclical} as the optimizer, with a default learning rate of 0.01, weight decay of 0.01, and momentum of 0.9. The learning rate is reduced at epochs 35 and 45 by a factor of 0.1, with a minimum learning rate clip of 1e$^{-7}$. The random seed is set as 0. The total batch size is set at 8, uniformly distributed among 4 $\times$ A6000 (48G) GPUs. In pseudo label generation, we follow previous works~\cite{you2022learning} to set $\alpha$ as 0.7 and $K$ as 20. Our code is based on OpenPCDet~\cite{openpcdet2020} and is implemented in PyTorch~\cite{paszke2019pytorch}.

\subsection{Main Results}

\begin{table}[t]
\centering
\caption{
Detection results on nuScenes. We report AP$_{BEV}$ and AP$_{3D}$ at $IoU=0.25$ for objects across various distances. The results are shown in AP$_{BEV}$ / AP$_{3D}$ format. $T=0$ is training from seed labels. $T=2$ and $T=10$ are the results for 2$th$ and 10$th$ round self-training, respectively. The supervised performance of model trained with ground-truth boxes is in the first row (Supervised). It is noticeable that the performance of LiSe significantly surpasses that of the state-of-the-art OYSTER~\cite{zhang2023towards} across all evaluated metrics.  $^*$: We present the results of our reimplementation, as official code for OYSTER is not available. Our reimplementation follows OYSTER settings, which conduct two rounds of self-training.
}
\label{tbl:nusc_025}
\renewcommand{\arraystretch}{1}
\small
\setlength{\tabcolsep}{7pt}
\begin{tabular}{lcccc}
\toprule
\textbf{Method} & \textbf{0-30m} & \textbf{30-50m} & \textbf{50-80m} & \textbf{0-80m} \\
\midrule
Supervised & 39.8 / 34.5 & 12.9 / 10.0 & 4.4 / 2.9 & 22.2 / 18.2 \\
\midrule
MODEST-PP ($T=0$) & 0.7 / 0.1 & 0.0 / 0.0 & 0.0 / 0.0 & 0.2 / 0.1 \\
MODEST-PP ($T=10$) & - & - & - & - \\
MODEST ($T=0$) & 16.5 / 12.5 & 1.3 / 0.8 & 0.3 / 0.1 & 7.0 / 5.0 \\
MODEST ($T=10$) & 24.8 / 17.1 & 5.5 / 1.4 & 1.5 / 0.3 & 11.8 / 6.6 \\
OYSTER ($T=0$) & 14.7 / 12.3 & 1.5 / 1.1 & 0.5 / 0.3 & 6.2 / 5.4 \\
OYSTER ($T=2$)$^*$ & 26.6 / 19.3 & 4.4 / 1.8 & 1.7 / 0.4 & 12.7 / 8.0 \\
\midrule
LiSe ($T=0$) & 5.8 / 4.7 & 0.6 / 0.2 & 0.3 / 0.2 & 2.1 / 1.8 \\
LiSe ($T=10$) & \textbf{35.0} / \textbf{24.0} & \textbf{11.4} / \textbf{4.4} & \textbf{4.8} / \textbf{1.3} & \textbf{19.8} / \textbf{11.4} \\
\bottomrule
\end{tabular}
\end{table}

We present nuScenes results in Table \ref{tbl:nusc_025} and observe that LiSe significantly outperforms all existing methods. In particular, compared to the state-of-the-art OYSTER, LiSe achieves an improvement of $+7.1$\% in AP$_{BEV}$ and $+3.4$\% in AP$_{3D}$ within the 0-80m range. In other distances, such as 0-30m, 30-50m, and 50-80m, LiSe consistently surpasses OYSTER, demonstrating a universally enhanced detection capability. The improvement validates the effectiveness of our proposed integration with 2D scenes, adaptive sampling strategy, and weak model aggregation in enhancing overall detection ability of model. It is also noteworthy that AP$_{BEV}$ of LiSe in the long range (50-80m) even exceeds that of fully supervised results. These results affirm that incorporating 2D scene understanding significantly augments the detection of distant and small objects.

\begin{table}[t]
\centering
\caption{
Comparison on Lyft. We observe that the proposed LiSe significantly surpasses MODEST across all evaluated metrics, especially in long range (50-80m). $T$ denotes self-training round.
}
\label{tbl:lyft_025} 
\renewcommand{\arraystretch}{1}
\small
\setlength{\tabcolsep}{5pt}
\begin{tabular}{lcccc}
\toprule
\textbf{Method} & \textbf{0-30m} & \textbf{30-50m} & \textbf{50-80m} & \textbf{0-80m} \\
\midrule
Supervised & 82.8 / 82.6 & 70.8 / 70.3 & 50.2 / 49.6 & 69.5 / 69.1 \\
\midrule
MODEST-PP ($T=0$) & 46.4 / 45.4 & 16.5 / 10.8 & 0.9 / 0.4 & 21.8 / 18.0 \\
MODEST-PP ($T=10$) & 49.9 / 49.3 & 32.3 / 27.0 & 3.5 / 1.4 & 30.9 / 27.3 \\
MODEST ($T=0$) & 65.7 / 63.0 & 41.4 / 36.0 & 8.9 / 5.7 & 42.5 / 37.9 \\
MODEST ($T=10$) & 73.8 / 71.3 & 62.8 / 60.3 & 27.0 / 24.8 & 57.3 / 55.1 \\
\midrule
LiSe ($T=0$) & 54.5 / 54.0 & 24.2 / 22.8 & 1.4 / 1.2 & 29.2 / 27.5 \\
LiSe ($T=10$) & \textbf{76.7} / \textbf{74.0} & \textbf{66.1} / \textbf{64.4} & \textbf{46.6} / \textbf{43.7} & \textbf{65.6} / \textbf{62.5} \\
\bottomrule
\end{tabular}
\end{table}

We further conduct experiments on Lyft, using the same hyper-parameters on nuScenes (see Table~\ref{tbl:lyft_025}).
We observe that the proposed LiSe surpasses the competitive MODEST across all evaluated metrics. More importantly, LiSe outperforms MODEST by $+19.4$\% in AP$_{BEV}$ and $+18.9$\% in AP$_{3D}$ in the long range (50-80m), which contributes most to overall improvement. These results validate the effectiveness and generalizability of our proposed method.

\subsection{Ablation Studies and Analyses}

\begin{table}[t]
\centering
\caption{
Ablation studies on the primary components of the proposed method, including integration with 2D scenes (3D and 2D), adaptive sampling strategy (ADS), and weak model aggregation (WMA).
}
\label{tbl:ablation_module}
\small
\renewcommand{\arraystretch}{1}
\setlength{\tabcolsep}{5pt}
\begin{tabular}{cccc|cccc}
\toprule
\textbf{3D} & \textbf{2D} & \textbf{ADS} & \textbf{WMA} & \textbf{0-30m} & \textbf{30-50m} & \textbf{50-80m} & \textbf{0-80m} \\
\midrule
\checkmark & & & & 24.8 / 17.1 & 5.5 / 1.4 & 1.5 / 0.3 & 11.8 / 6.6 \\
& \checkmark & & & 31.8 / 14.0 & 3.8 / 0.8 & 0.6 / 0.0 & 12.4 / 4.7 \\
\checkmark & \checkmark & & & 31.4 / 19.9 & 8.3 / 3.1 & 3.4 / 0.9 & 16.2 / 9.1 \\
\checkmark & \checkmark & \checkmark & & 32.8 / 22.3 & 11.1 / 3.8 & 3.9 / 0.9 & 18.4 / 10.2 \\
\checkmark & \checkmark & & \checkmark & 34.3 / 23.3 & 10.0 / 4.1 & 4.0 / 1.3 & 18.5 / 11.0 \\
\checkmark & \checkmark & \checkmark & \checkmark & \textbf{35.0} / \textbf{24.0} & \textbf{11.4} / \textbf{4.4} & \textbf{4.8} / \textbf{1.3} & \textbf{19.8} / \textbf{11.4} \\
\bottomrule
\end{tabular}
\end{table}

\begin{table}[t]
\caption{
Ablation studies on integrating image-based 3D boxes according to the distance. We incorporate image-based 3D boxes with a distance greater than 5, 10, and 15 meters (>5m, >10m, >15m). The term ``All'' refers to the use of all image-based boxes. We find that >10m yields the optimal results. This threshold balancedly integrates the advantage of 2D scenes and avoids conflict with LiDAR-based 3D boxes in near range.
}
\centering
\small
\setlength{\tabcolsep}{5pt}
\begin{tabular}{cc|cccc}
\toprule
\textbf{3D} & \textbf{2D} & \textbf{0-30m} & \textbf{30-50m} & \textbf{50-80m} & \textbf{0-80m} \\
\midrule
All & All & 29.3 / 19.8 & 4.7 / 2.3 & 2.4 / 0.5 & 13.8 / 8.0 \\
All & >5m & 30.7 / 19.8 & \textbf{8.6} / \textbf{3.3} & 3.1 / 0.7 & 15.4 / 8.9 \\
All & >10m & \textbf{31.4} / \textbf{19.9} & 8.3 / 3.1 & \textbf{3.4} / \textbf{0.9} & \textbf{16.2} / \textbf{9.1} \\
All & >15m & 30.2 / 20.6 & 5.7 / 2.8 & 2.1 / 0.4 & 14.3 / 8.9 \\
\bottomrule
\end{tabular}
\label{tbl:ablation_integration_detail}
\end{table}

\begin{table}[t]
\centering
\caption{
Adaptive sampling strategy based on the volume of objects (Volume) or the distance of objects (Distance). We could observe that volume-based strategy facilitates the distant objects in 30-50m, while distance-focused sampling improves box detection, with remarkable improvement in median distance. After combining two factors into consideration, we arrive at a balanced strategy for all ranges.
}
\small
\setlength{\tabcolsep}{5pt}
\label{tbl:ablation_asm}
\begin{tabular}{cc|cccc}
\toprule
\textbf{Volume} & \textbf{Distance} & \textbf{0-30m} & \textbf{30-50m} & \textbf{50-80m} & \textbf{0-80m} \\
\midrule
& & 31.4 / 19.9 & 8.3 / 3.1 & 3.4 / 0.9 & 16.2 / 9.1 \\
\checkmark & & 31.5 / 20.1 & 10.1 / 3.0 & 3.4 / 0.6 & 17.1 / 8.6 \\
& \checkmark & \textbf{35.0} / 22.2 & 10.5 / \textbf{3.8} & 3.2 / 0.7 & 18.2 / 9.6 \\
\checkmark & \checkmark & 32.8 / \textbf{22.3} & \textbf{11.1} / \textbf{3.8} & \textbf{3.9} / \textbf{0.9} & \textbf{18.4} / \textbf{10.2} \\
\bottomrule
\end{tabular}
\end{table}

\begin{table}[t]
\centering
\small
\caption{
Ablation study of starting round $T_{s}$ and the aggregation coefficient $\lambda$ selection during the weak model aggregation. (1) We fix $\lambda$ and study $T_{s}$. We observe that initiating the aggregation process at a later round, when the performance of model is higher and fluctuating, yields better results. (2) If we fix $T_s$ as 6, we can see the large $\lambda$ with slow update speed has achieved the best results in all different ranges, which stabilizes the prediction result.
}
\label{tbl:ablation_wma}
\setlength{\tabcolsep}{5pt}
\begin{tabular}{cc|cccc}
\toprule
\textbf{$\bm{T_{s}}$} & $\bm{\lambda}$ & \textbf{0-30m} & \textbf{30-50m} & \textbf{50-80m} & \textbf{0-80m} \\
\midrule
3 & 0.999 & 31.7 / 21.5	& 8.7 / 4.2	& 2.7 / 0.7	& 16.3 / 10.2 \\
6 & 0.999 & 35.4 / 21.3 & 10.0 / 4.0 & 3.2 / 0.6 & 18.5 / 10.1 \\
8 & 0.999 & \textbf{34.3} / \textbf{23.3} & \textbf{10.0} / \textbf{4.1} & \textbf{4.0} / \textbf{1.3} & \textbf{18.5} / \textbf{11.0} \\
\midrule
6 & 0.999 & \textbf{35.4} / \textbf{21.3} & \textbf{10.0} / \textbf{4.0} & \textbf{3.2} / \textbf{0.6} & \textbf{18.5} / \textbf{10.1} \\
6 & 0.99 & 31.9 / 20.7 & 8.8 / 3.1 & 2.9 / 0.5 & 16.5 / 9.0 \\
6 & 0.9 & 31.4 / 21.0 & 7.4 / 3.3 & 2.3 / 0.5 & 16.0 / 9.5 \\
\bottomrule
\end{tabular}
\end{table}

\noindent \textbf{Effect of Integration with 2D Scenes.} Comparing the first three rows in Table~\ref{tbl:ablation_module}, we observe that integrating 2D scenes into 3D-based pseudo boxes yields the best overall performance across both AP$_{BEV}$ and AP$_{3D}$. The significant improvement over LiDAR-based methods highlights the unique advantages of 2D scenes in detecting distant and small objects. We further examine the way to integrate 2D scenes (see Table~\ref{tbl:ablation_integration_detail}). Specifically, we only incorporate 3D boxes from images with distance over 5m, 10m, and 15m. Table~\ref{tbl:ablation_integration_detail} indicates by integrating image-based boxes with distance over 10m achieves best performance. Such results also suggest that image-based boxes and LiDAR-based boxes conflict with each other in range 0-10 meters. Consideration of object distance in integration avoids such conflicts, and make two modalities complementary.

\begin{figure*}[t]
    \centering
    \includegraphics[width=\linewidth]{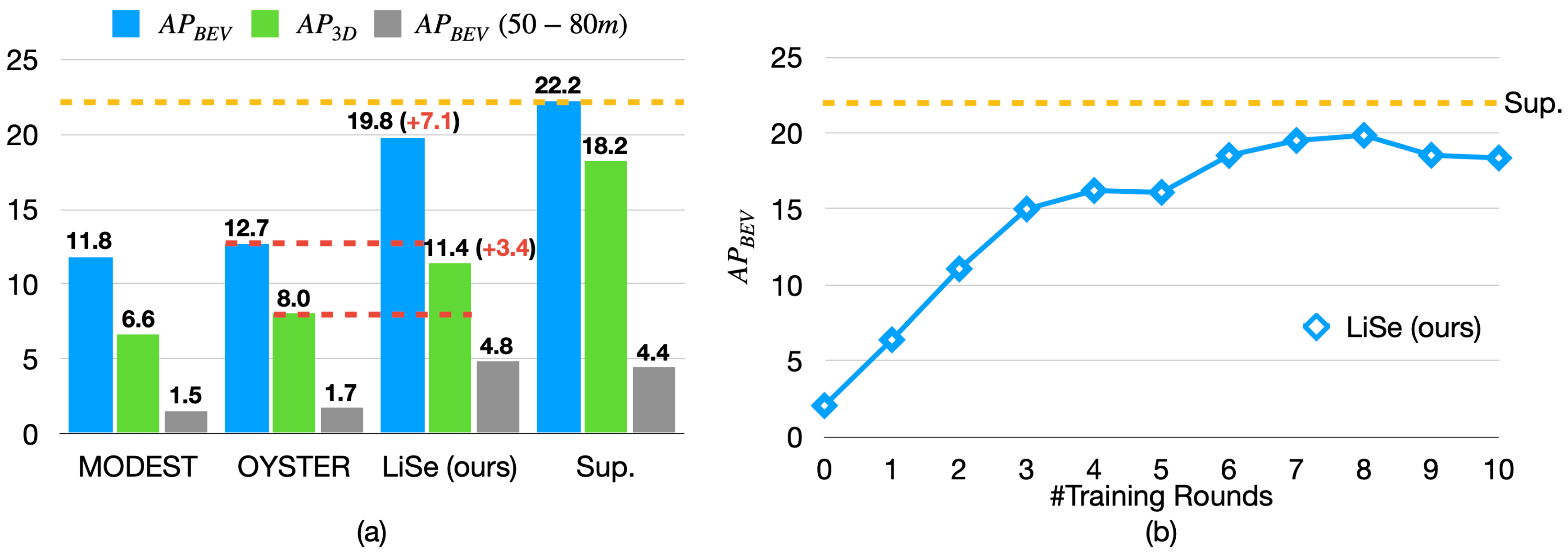}
    \caption{
    Statistical analysis of performance of different models. (a) Visualization comparison of the performances of various methods. This comparison shows superiority of LiSe over purely LiDAR-based methods. (b) Visualization of performance changes throughout the training process. The trend shows combination of adaptive sampling strategy with weak model aggregation ensures a stable and effective training process.
    }
    \label{fig:analysis}
\end{figure*}

\noindent \textbf{Effect of Adaptive Sampling Strategy.} Comparing rows 3 and 4 in Table~\ref{tbl:ablation_module}, it shows that adaptive sampling further enhances model performance, particularly in the 30-50m and 50-80m ranges, validating its effectiveness in enhancing long-range detection capabilities. We also conduct a comprehensive ablation study to evaluate the design of adaptive sampling strategy. According to Table~\ref{tbl:ablation_asm}, the inclusion of a single metric, such as volume or distance, significantly enhances model performance. Notably, when incorporating a distance-based metric, both AP$_{BEV}$ and AP$_{3D}$ in the 30-50m range exceed the performance achieved with a volume-based metric alone. The improvement underscores the effectiveness of distance-based adaptive sampling in improving long-range detection capabilities. The combination of both volume-based and distance-based metrics yields the best performance, demonstrating that combined metric can more comprehensively addresses overlooking on long-tailed object groups. These two metrics work in a complementary manner, further enhancing overall model efficacy.

\noindent \textbf{Effect of Weak Model Aggregation.} Comparing rows 3 and 5 in Table~\ref{tbl:ablation_module}, we observe that weak model aggregation alone enhances model performance. This improvement demonstrates effectiveness of weak model aggregation in creating a more robust model. Furthermore, when comparing rows 4, 5, and 6 in Table~\ref{tbl:ablation_module}, we find combination of adaptive sampling strategy and weak model aggregation yields the best performance. This validates crucial role of interaction between adaptive sampling strategy and weak model aggregation. We also examine impact of start round $T_{s}$ and aggregation coefficient $\lambda$ in Table~\ref{tbl:ablation_wma}. Initially, we fix $\lambda$ at 0.999 and vary start round in 3, 6, and 8. The findings suggest that initiating aggregation process at a later round, when performance of model is higher and fluctuating, yields better results. During this period, models generally exhibit good performance with distinct strengths, making it an opportune time for weak model aggregation. Additionally, we fix start round and vary aggregation coefficient $\lambda$ in 0.999, 0.99, and 0.9. Observations indicate larger coefficient, such as 0.999, leads to best performance. The ablation implies aggregation process benefits from being smoother and progressing in smaller steps.

\noindent \textbf{Statistical analyses.} We present statistical analyses of performance of different models in Figure~\ref{fig:analysis}. In Figure~\ref{fig:analysis}(a), LiSe significantly outperforms state-of-the-art model, OYSTER~\cite{zhang2023towards}. Notably, AP$_{BEV}$ of LiSe at long ranges (50-80m) surpasses that of fully-supervised results. It verifies our integration with 2D scenes effectively enhances capability of model to detect distant objects. In Figure~\ref{fig:analysis}(b), we can observe LiSe starts at a low point in initial training round, yet consistently achieves improved performance during self-paced training process.

\begin{figure*}[t]
    \centering
    \includegraphics[width=\linewidth]{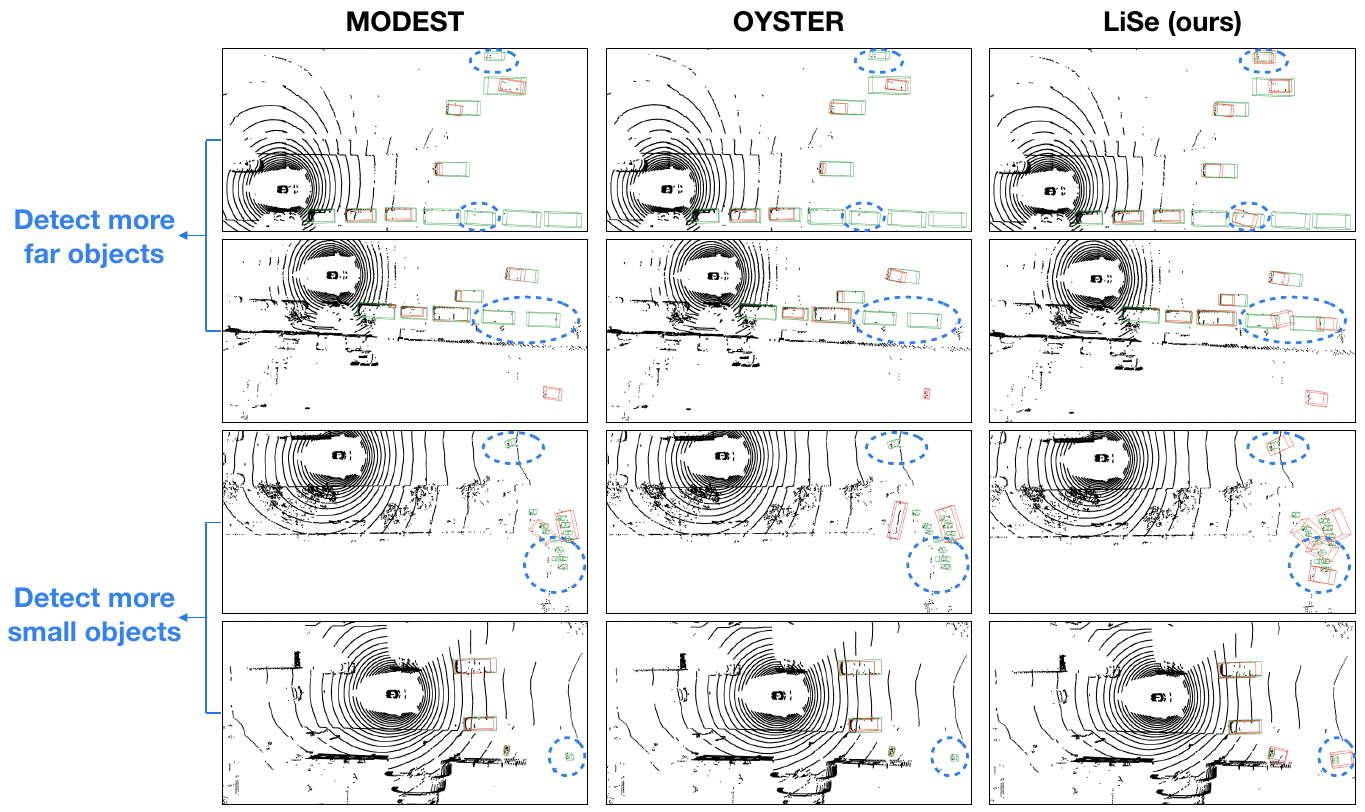}
    \caption{
    Visualization comparison between MODEST~\cite{you2022learning}, OYSTER~\cite{zhang2023towards}, LiSe (ours), and ground truth boxes. The overall results indicate LiSe is superior in detecting distant and small objects. \textcolor{my_green}{Green} boxes represent ground truth labels, \textcolor{my_red}{red} boxes indicate predictions and \textcolor{my_blue}{blue} circles highlight differences in predictions.
    }
    \label{fig:qualitative}
\end{figure*}

\noindent \textbf{Visualization.}  We present qualitative analysis in Figure \ref{fig:qualitative}. From rows 1 and 2, we observe LiSe excels at detecting distant objects, even when there are very limited points captured. Results in rows 3 and 4 indicate our LiSe framework performs significantly better than MODEST and OYSTER in detecting small objects. These results validate our obtained model is robust in detecting potentially existing objects, especially for challenging distant and small objects.

\section{Conclusion}

In this paper, we introduce a framework \textbf{LiSe} for unsupervised 3D detection. We propose integration with 2D scenes to improve detection ability in distant and small objects. In self-paced learning process, we further propose adaptive sampling strategy to continuously improve perception ability in challenging samples. Additionally, we introduce weak model aggregation, combining models trained under different distributions into a final, robust model. Extensive experiments affirm superior detection ability of our method. The comprehensive ablation studies and qualitative analyses also validate effectiveness of each proposed module. We hope our work will contribute to the fusion between 2D and 3D data for unsupervised 3D object detection and inspire future work in related fields.

\section*{Acknowledgement}
The paper is supported by Start-up Research Grant at the University of Macau (SRG2024-00002-FST).

%
%
\bibliographystyle{splncs04}
\bibliography{main}
\end{document}